\documentclass[10pt, a4paper]{article}
\usepackage{amsmath}
\usepackage{graphicx}
\usepackage{pgfplotstable}
\usepackage{tabularx}  
\usepackage{booktabs}
\usepackage[table]{xcolor} 
\usepackage{makecell,multirow}

\usepackage{lrec2026} 

\title{Estonian Native Large Language Model Benchmark}

\name{Helena Grete Lillepalu, Tanel Alumäe} 

\address{Department of Software Science, Tallinn University of Technology, Estonia\\
         tanel.alumae@taltech.ee}

\abstract{
The availability of LLM benchmarks for the Estonian language is limited, and a comprehensive evaluation comparing the performance of different LLMs on Estonian tasks has yet to be conducted. We introduce a new benchmark for evaluating LLMs in Estonian, based on seven diverse datasets. These datasets assess general and domain-specific knowledge, understanding of Estonian grammar and vocabulary, summarization abilities, contextual comprehension, and more. The datasets are all generated from native Estonian sources without using machine translation.
We compare the performance of base models, instruction-tuned open-source models, and commercial models.
Our evaluation includes 6 base models and 26 instruction-tuned models. To assess the results, we employ both human evaluation and LLM-as-a-judge methods. Human evaluation scores showed moderate to high correlation with benchmark evaluations, depending on the dataset. Claude 3.7 Sonnet, used as an LLM judge, demonstrated strong alignment with human ratings, indicating that top-performing LLMs can effectively support the evaluation of Estonian-language models.
 \\ \newline  }

\begin{document}

\maketitleabstract

\section{Introduction}
Performance benchmarks play an important role in evaluating the capabilities of large language models (LLMs). These standardized tests enable researchers to assess various aspects of a model's language proficiency, such as text comprehension, reasoning ability, and generalization skills. For widely spoken languages, a variety of performance benchmarks have been developed to measure model performance across different tasks \cite{hendrycks2021measuringmassivemultitasklanguage, wang2019gluemultitaskbenchmarkanalysis, zellers2019hellaswagmachinereallyfinish}.

While these benchmarks have provided valuable insights into the capabilities of LLMs in languages with large speaker populations, similar comprehensive evaluation frameworks are lacking for many languages with smaller speaker bases, including Estonian, a Uralic language with around one million native speakers.

We present a benchmark suite designed to evaluate the capabilities of LLMs on a variety of tasks in the Estonian language. The tasks include factual recall (exam and trivia questions), linguistic accuracy (word meanings, grammatical corrections, inflection), summarization, and structured information extraction from long texts (speaker name extraction). Together, these tasks offer a framework for assessing both the general and language-specific performance of LLMs in Estonian. 

We evaluate a range of open- and closed-source base and instruction-tuned models. In total, we assess 6 base models and 26 instruction-tuned (chat) models.  The instruction-tuned set includes many open-source models of different sizes, and 7 commercial models. Base models are evaluated in a 5-shot setting, while instruction-tuned models are evaluated in a zero-shot setting. 
We compare individual models across all tasks and also perform a broader comparison based on model types, specifically base models, open-source instruction-tuned models, and commercial models. 

To evaluate the results, we combine human judgment with the "LLM-as-a-judge" methodology \cite{zheng2023judgingllmasajudgemtbenchchatbot}. We create a set of localized open-ended questions that users might typically ask from LLMs. A custom web interface is developed to present one question alongside responses from two anonymous models, allowing users to choose the better answer, or indicate if both are equally good. For automated evaluation, we select Claude Sonnet as the LLM judge due to its strong performance in benchmark tests, and task it with evaluating the answers using the same criteria as the human evaluators. 

To summarize, our contributions are:
\begin{itemize}
    \item We present a benchmark suite for evaluating LLMs in Estonian, based on four new datasets and three existing ones, along with \texttt{lm-evaluation-harness}~\cite{eval-harness} configurations for automated testing.
    \item We evaluate a wide range of open and proprietary LLMs on the proposed benchmark.
    \item We conduct a human evaluation of chat models on a set of open-ended questions.
    \item We show that Claude 3.7 Sonnet can reliably judge open-ended responses with agreement comparable to human raters.
    \item We analyze which benchmark tasks correlate best with human evaluations.
\end{itemize}

\section{Related work}
\subsection{Types of LM benchmarks} 
As LLMs have become more advanced, a variety of benchmarks have been developed to evaluate their capabilities across different dimensions. General-purpose benchmarks such as GLUE and SuperGLUE are widely used to assess language understanding and reasoning skills \cite{wang2019gluemultitaskbenchmarkanalysis, wang2020supergluestickierbenchmarkgeneralpurpose}, while others focus specifically on reading comprehension through question-answering tasks \cite{rajpurkar2016squad100000questionsmachine, lai2017racelargescalereadingcomprehension}. More specialized benchmarks have also emerged. For example, MMLU evaluates subject-specific knowledge across 57 academic and professional domains using multiple-choice questions \cite{hendrycks2021measuringmassivemultitasklanguage}, and HellaSwag is designed to assess common-sense reasoning \cite{zellers2019hellaswagmachinereallyfinish}. In addition, several benchmarks target code generation by requiring models to write syntactically and functionally correct programs \cite{peng2024humanevalxlmultilingualcodegeneration, puri2021codenetlargescaleaicode}. Bias detection has also become a focal point, with benchmarks developed to measure biases in text generation across various social and demographic dimensions \cite{Dhamala_2021}. Collectively, these benchmarks reflect the growing effort to comprehensively evaluate LLMs across a wide range of tasks and domains. 

\subsection{Benchmarking for underrepresented languages}
While numerous high-quality benchmarks have been developed for high-resource languages, comparable resources for low-resource languages remain scarce. Languages such as Estonian face a notable gap in evaluation frameworks, which poses challenges for researchers and developers aiming to assess or fine-tune language models for these languages. Nevertheless, recent initiatives have begun to address this gap by developing evaluation suites tailored to low-resource languages. These efforts typically span a variety of tasks, ranging from general language understanding to more specialized applications such as reading comprehension and question answering \cite{etxaniz2024latxaopenlanguagemodel, fajcik2024benczechmarkczechcentricmultitask}. In some cases, the focus has been on creating localized analogues of well-known English benchmarks, such as MMLU, adapted for specific low-resource languages \cite{yüksel2024turkishmmlumeasuringmassivemultitask, koto2024arabicmmluassessingmassivemultitask}.

A common approach to creating LLM benchmarks for low-resource languages is through manual or automatic translation.  For example, the EU20 benchmark \cite{thellmann2024towards} contains benchmarks originally in the English language, such as MMLU, HellaSwag, TruthfulQA, machine-translated into 20 European languages. Similarly, the M-MMLU benchmark~\cite{hendrycks2021measuringmassivemultitasklanguage} provides professional human translations of the MMLU test set into 14 languages.

However, translated benchmarks introduce two major issues: translation noise and reduced cultural relevance.
For example, Global-MMLU \cite{singh2025global} showed that success on MMLU relies heavily on Western-centric concepts, with 28\% of questions requiring culturally sensitive knowledge and 84.9\% of geographically focused questions centered on North America or Europe.
The lower the resource level of the target language, the more likely it is that machine translation will introduce errors, as translation quality tends to be poorer for such languages. This becomes particularly problematic when comparing model performance across multiple languages.
Even with high-quality human translations, benchmarks may still fall short of providing reliable evaluations unless they are properly localized. Research has shown that localized benchmarks that are adapted to the culture and context of the target language correlate better with human preferences than those that are only translated. 
\citet{wu2025bitterlessonlearned2000} demonstrated that localized benchmarks (such as CMMLU  \cite{li-etal-2024-cmmlu} for Chinese) exhibit a correlation of 0.68 with human judgments, whereas translated versions of the same benchmarks fall significantly lower, reaching only 0.47 to 0.49. Similarly, \citet{son-etal-2025-kmmlu} introduce KMMLU for Korean and find its Spearman correlation with human model-ranking (0.94) greatly exceeds that of a translated MMLU (0.86).
This provides a clear statistical confirmation that simply translating evaluation sets is insufficient for capturing true human preference.

\section{Benchmarks}
To evaluate LLM performance in Estonian, we constructed a benchmark suite based on seven datasets, four created manually and three adapted from existing resources.

To convert a dataset of questions and reference answers into a LLM benchmark, we defined \texttt{lm-evaluation-harness} ``tasks''. Each task specifies how a dataset item is converted into the LLM prompt, the method for extracting answers from model outputs, and the implementation of the evaluation metric. 

For each individual benchmark, we implemented separate task definitions for base LLMs (using a 5-shot setting for most benchmarks) and for instruction-tuned LLMs (using a zero-shot setting). This distinction reflects the typical usage of these model types: base models generally require a few in-context examples to reliably follow task instructions, whereas instruction-tuned chat models are explicitly trained to follow natural language instructions and therefore are appropriately evaluated in a zero-shot setting.

\textbf{Estonian National Exam benchmark}:
This benchmark is based on the Estonian National Exam dataset (links to datasets are provided at the end of the paper) that contains official exam questions from Estonian secondary and high school exams (2003–2024). It includes 1614 multiple-choice questions across seven subjects: Estonian as a second language, physics, chemistry, biology, geography, history, and civic studies.

The Estonian as a second language section contains 484 questions (246 secondary school, 238 middle school), primarily focused on gap-filling and reading comprehension, testing grammatical accuracy and text understanding. Other subjects contain between 108 and 310 questions each, targeting factual knowledge and reasoning skills. The question formats include simple factual queries, gap-filling tasks, and true/false statements, all presented as multiple-choice questions with 2 to 15 answer options.

The dataset was compiled based on PDFs obtained from the state's education board. The PDFs were converted to text, edited for conversion errors and manually restructured into triplets of question, answer choices and the reference answer. An example question from this benchmark is shown in Table \ref{tab:example-exams}. 

\begin{table}[tb]
\centering
\scriptsize
\begin{tabular}{p{6cm}c}
\toprule
Prompt & Answer \\
\midrule
Sa oled ekspert ning vastad küsimustele korrektselt. Sulle antakse küsimus koos vastusevariantidega. Anna vastus kujul: “Vastus: X”, kus X on kas A, B või C. Ära kirjuta midagi muud! Vasta eesti keeles!\\ Küsimus: Ühemunakaksikud on...\\ Vastusevariandid:\\ A: ühest soost\\ B: eri soost\\ C: nii ühest kui ka eri soost.                                        & A \\ \midrule
You are an expert and answer questions correctly. You are given a question with answer options. Provide the answer in the form: “Answer: X”, where X is either A, B, or C. Do not write anything else! Answer in Estonian!\\ Question: Identical twins are...\\ Answer options:\\ A: of the same sex\\ B: of different sexes\\ C: both of the same and different sexes. & A \\
\bottomrule
\end{tabular}
\caption{Question-answer pair from the Exams benchmark, together with its English translation. Question is already formatted as a prompt to an instruction-tuned LLM.}
\label{tab:example-exams}
\end{table}

\textbf{Trivia benchmark} is based on the Trivia dataset that consists of 800 multiple-choice questions from the board game ``\textit{Eesti mälumäng}'', covering history, science, culture, geography, sports, and more. Most questions are highly specific to Estonia or Estonian culture. The task evaluates general and Estonia-specific knowledge. 

\begin{figure}[tb]
    \centering
    \includegraphics[width=1\linewidth]{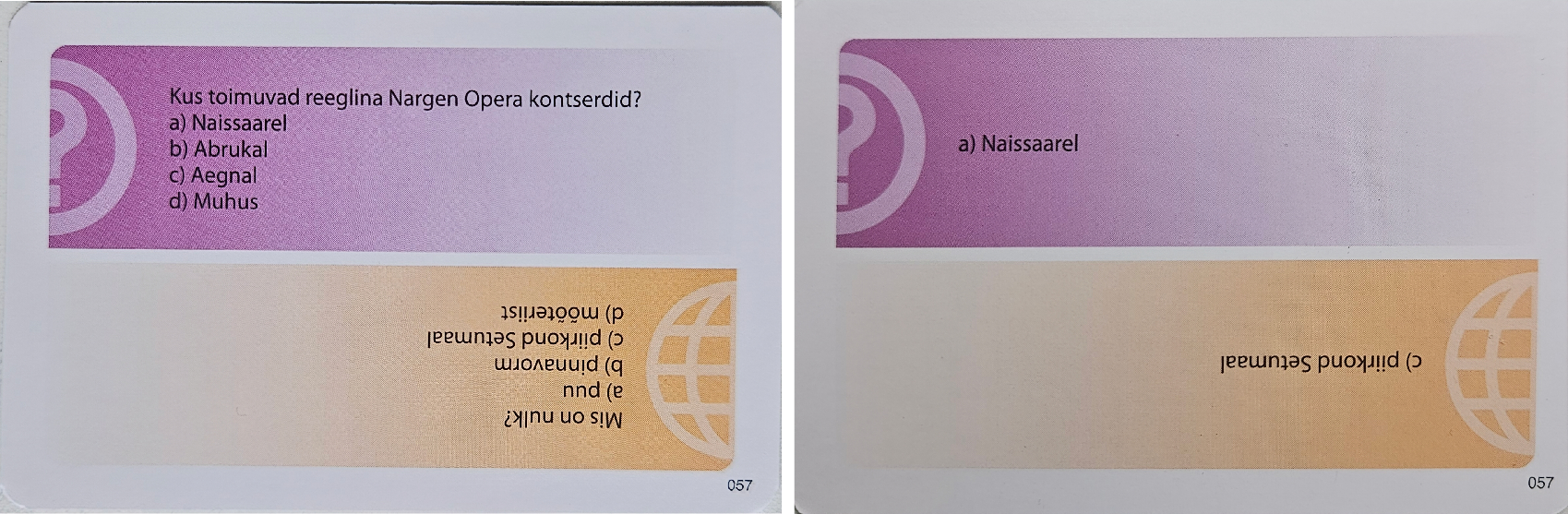}
    \caption{A game card from the quiz boardgame that was the source for the Trivia dataset.}
    \label{fig:trivia}
\end{figure}

The dataset was compiled by scanning 400 individual double-sided game cards, each with two questions (one to be read up-side down), and with answers written on the opposite side of the card (see Figure \ref{fig:trivia}). Scanned cards were OCR-ed, restructured using an LLM and post-edited for errors. An example sample prompt from this benchmark is shown in Table \ref{tab:example-trivia}. 

\begin{table}[tb]
\centering
\scriptsize
\begin{tabular}{p{6cm}p{1cm}}
\toprule
Prompt & Answer \\
\midrule
Sa oled ekspert ning vastad küsimustele korrektselt. Sulle antakse küsimus koos vastusevariantidega. Anna vastus kujul: ``Vastus: X'', kus X on kas A, B, C või D. Ära kirjuta midagi muud! Vasta eesti keeles! Küsimus: Kus toimuvad reeglina Nargen Opera kontserdid? Vastusevariandid: A: Naissaarel B: Abrukal C: Aegnal D: Muhus & A \\
\midrule
You are an expert and answer questions correctly. You are given a question with answer options. Provide the answer in the form: ``Vastus: X'', where X is either A, B, C, or D. Do not write anything else! Answer in Estonian!
Question: Where are Nargen Opera concerts usually held?
Answer options:
A: Naissaar
B: Abruka
C: Aegna
D: Muhu & A \\
\bottomrule
\end{tabular}
\caption{Question-answer pair from the Trivia benchmark, together with translation.}
\label{tab:example-trivia}
\end{table}

\textbf{Declension benchmark} uses the Declension dataset and
evaluates LLMs’ ability to generate correct grammatical case forms in Estonian -- a morphologically rich language with 14 noun cases. The task consists of 200 adjective-noun pairs, each combined with 7 target case-number combinations (e.g., plural partitive, singular illative), resulting in 1400 total samples. Gold standard forms were generated using the Filosoft Estonian morphological synthesizer \cite{kaalep2001complete}. An example question from this benchmark is shown in Table \ref{tab:example-inflection}. 

\begin{table}[tb]
\centering
\scriptsize
\begin{tabular}{p{6cm}p{1cm}}
\toprule
Prompt & Answer \\
\midrule
Sa oled eesti keele ekspert. Sulle antakse sõnapaar ainsuse nimetavas käändes ja sinu ülesandeks on käänata see etteantud käändesse. Jälgi, kas käänet on küsitud ainsuses või mitmuses. Anna vastus kujul: "Vastus: \{käänatud sõnapaar\}". 
Sõnapaar ainsuse nimetavas käändes on: kollane päevalill.
Kääna see käändesse: ainsuse omastav & kollase päevalille \\
\midrule
You are an expert in the Estonian language. You are given a word pair in the singular nominative case, and your task is to inflect it into the given case. Pay attention to whether the requested case is singular or plural. Provide the answer in the format: "Answer: \{inflected word pair\}".
The word pair in the singular nominative case is: kollane päevalill.
Inflect it into the case: singular genitive & kollase päevalille \\
\bottomrule
\end{tabular}
\caption{Question-answer pair from the Declension benchmark, together with translation.}
\label{tab:example-inflection}
\end{table}

\textbf{Word Meaning benchmark} tests lexical understanding by prompting the model with a word definition and asking it to generate the correct term. The underlying dataset 
is derived from Ekilex, Estonian word and term base system of the Institute of the Estonian Language
  \cite{tavast2023ekilex}. The full set contains over 54,000 entries, from which a random sample of 1000 items was used in the benchmark. Synonyms obtained via Estonian WordNet \cite{orav2018estonian} were accepted as correct answers. A sample item from this benchmark is shown in Table \ref{tab:example-words}.

\begin{table}[tb]
\centering
\scriptsize
\begin{tabular}{p{5.5cm}p{1.5cm}}
\toprule
Prompt & Answer \\
\midrule
Sa oled sõnade teadmise ekspert. Sulle antakse sõna definitsioon ning sina pead tagastama sõna, mida see definitsioon kirjeldab. Anna vastus kujul: ``Vastus: \{definitsioonile vastav sõna\}''. Vasta ainult vastusega. Sõna definitsioon(id): ['suuri kulusid, väljaminekuid nõudev', '(esemete kohta:) hinnalt kallis, kalli, kõrge hinnaga'] &  kulukas, kallis, kallivõitu, kõrgehinnaline, soolane, krõbe, kallis, kallihinnaline, hinnaline \\\\
\midrule
You are an expert in knowledge about words. You are given a word definition, and you must return the word that the definition describes. Provide the answer in the format: ``Answer: {word corresponding to the definition}''. Respond with the answer only.
Word definition(s): ['requiring large expenses or costs', '(of objects:) with a high price'] & expensive, high-priced, premium, pricey \\
\bottomrule
\end{tabular}
\caption{Question-answer pair from the Word Meaning benchmark, together with translation.}
\label{tab:example-words}
\end{table}

\textbf{Grammar Correction benchmark} is based on Estonian grammar correction dataset which in turn are simplified versions of the EstGEC corpus. They contain original and corrected sentences written by native and non-native Estonian students (446 and 3285 sentences respectively). The model is tasked with correcting grammatical errors if present. Evaluation uses both exact-match (binary) accuracy and Levenshtein-based similarity scores.   A sample question-answer pair from this benchmark is shown in Table \ref{tab:example-grammar}.

\begin{table}[tb]
\centering
\scriptsize
\begin{tabular}{p{5.5cm}p{1.5cm}}
\toprule
Prompt & Answer \\
\midrule
Sa oled eesti keele ekspert. Sulle antakse ette eestikeelsed laused, kus võivad esineda grammatikavead. Tagasta korrektne lause, kus vead on parandatud. Juhul kui etteantud lause on juba õige, siis tagastagi see sama lause. Vasta ainult lausega! Originaalne: Kellena küll saaksin, kui ei oskaks lugeda? Mitte täiusliku inimesena. & Kelleks küll saaksin, kui ei oskaks lugeda? Mitte täiuslikuks inimeseks. \\
\midrule
You are an expert in the Estonian language. You are given Estonian sentences that may contain grammatical errors. Return the correct sentence with the errors fixed. If the given sentence is already correct, return it as is. Respond with the sentence only!
Original: Whom could I possibly become if I couldn’t read? Not as a perfect person. &  Who could I possibly become if I couldn’t read? Not a perfect person. \\
\bottomrule
\end{tabular}
\caption{Question-answer pair from the Grammar correction benchmark, together with translation.}
\label{tab:example-grammar}
\end{table}

\textbf{News Summarization benchmark} uses the ERRNews dataset that contains transcriptions of Estonian public broadcaster (ERR) radio news stories paired with human-written short summaries \cite{harm2022abstractive}. Transcripts are generated using an automatic speech recognition (ASR) system. From over 10,000 records, a test split of 523 items was used. Evaluation uses ROUGE-L metrics. A sample response-answer pair from this benchmark is shown in Table \ref{tab:example-news}.

\begin{table}[tb]
\centering
\scriptsize
\begin{tabular}{p{5.0cm}p{2.0cm}}
\toprule
Prompt & Answer \\
\midrule
Sa oled tehisintellekti ekspert eesti keele alal. Su ülesanne on analüüsida uudislugusid ja genereerida ühe või kahe lausega lühike kokkuvõte, mis kirjeldab uudisloo sisu. Ära unusta alati vastata eesti keeles. Vasta ainult kokkuvõttega! Uudislugu: Eesti Energia hinnangul oli auvere elektrijaama ehitus keerukas pikk ja pingeline protsess. Ehituslepingu sõlmimisest kuni jaama käivitamiseni jõuti nelja aastaga ... & Eesti Energia võttis ehitajalt üle kauaoodatud Auvere elektrijaama. Eesti moodsaim põlevkivijaam suudab katta umbes neljandiku riigi elektrivajadusest. \\
\midrule
You are an artificial intelligence expert in the Estonian language. Your task is to analyze news stories and generate a short one- or two-sentence summary that describes the content of the news story. Always remember to respond in Estonian. Respond with the summary only! News story: According to Eesti Energia, the construction of the Auvere power plant was a complex, long, and demanding process. It took four years from signing the construction contract to the commissioning of the plant... & Eesti Energia took over the long-awaited Auvere power plant from the builder. Estonia’s most modern oil shale power plant can cover about a quarter of the country’s electricity needs. \\
\bottomrule
\end{tabular}
\caption{Sample prompt and expected answer from the News Summarization benchmark, together with translation.}
\label{tab:example-news}
\end{table}

\textbf{Speaker Name Extraction benchmark} is based on a dataset consisting of speaker-attributed transcripts from the main evening radio news program of the Estonian Public Broadcasting (\textit{Päevakaja}), along with the corresponding list of speakers. The speaker list includes both news anchors and journalists, who typically introduce themselves, as well as interviewees introduced by the moderators. The reference lists of speakers were manually compiled by human editors. 

The transcripts were generated using an ASR system with a word error rate of around 8\%, and each speaking turn is prefixed with a speaker code obtained from a speaker diarization system. The LLM’s task is to identify all speakers occurring  in the show. This benchmark evaluates the model’s ability to follow instructions, understand text, and extract structured information. A fuzzy heuristic algorithm is employed to divide the list of names generated by the LLM into individual names, while permitting some flexibility in how apostrophes and separators around the names are handled.
Performance is measured using precision, recall, and F1-score (the main metric), using exact speaker name match. 
A sample item from this benchmark is shown in Table \ref{tab:example-speakers}.

\begin{table}[h]
\centering
\scriptsize
\begin{tabular}{p{5.0cm}p{2.0cm}}
\toprule
Prompt & Answer \\
\midrule
Sa oled ekspert Eesti avaliku elu tegelaste alal. Sulle antakse uudiste- või vestlussaate automaatne transkriptsioon, koos kõneleja koodidega. Proovi arvata, millised isikud saates kõnelevad, nii reporterid kui ka intervjueeritavad. Veendu et sa vastad inimestega, kes kõnelevad, mitte inimestega, kellest räägitakse. Väljasta tulemus JSONi abil. JSON väljundi formaadi näide: ["nimi1", "nimi2"].  Ära liialt riski, meile on täpsus olulisem kui saagis. Nimed võivad olla valesti transkribeeritud, kasuta oma taustateadmisi, et
 neid vajadusel korrigeerida. Kindlasti anna vastuseks ainult üks JSON list ja mitte midagi muud! & \multirow{2}{*}{\parbox[t]{2cm}{"Margitta Otsmaa", "Jaak Aab", "Alvar Soesoo", ...}}\\
S1: Tere õhtust, kell on kuus ning Päevakaja võtab kokku tänasest olulisema meil ja maailmas, mina olen toimetaja Margitta otsmaa... & \\
\midrule
You are an expert on Estonian public figures. You are given an automatic transcript of a news or talk show, along with speaker codes. Try to identify which individuals are speaking in the program, including both reporters and interviewees. Make sure you name the people who are speaking, not those who are being talked about. Output the result in JSON format. Example of JSON output format: ["name1", "name2"]. Don’t take unnecessary risks — accuracy is more important than recall for us. The names may be misspelled in the transcript; use your background knowledge to correct them if necessary. Be sure to respond with only one JSON list and nothing else! & \multirow{2}{*}{\parbox[t]{2cm}{"Margitta Otsmaa", "Jaak Aab", "Alvar Soesoo", ...}}\\
S1: Good evening, it’s six o’clock and Päevakaja brings you a summary of the most important news from Estonia and the world; I’m editor Margitta otsmaa... \\
\bottomrule
\end{tabular}
\caption{Sample prompt and expected answer from the Speaker Name Extraction benchmark, together with translation. Note how the name of the editor \textit{Margitta Otsmaa} has a casing error in the automatic transcript.}
\label{tab:example-speakers}
\end{table}

\section{Results}

We evaluated a total of 6 base models and 26 chat models on the seven benchmark tasks described above. For all benchmarks, we used metrics (such as accuracy, ROUGE and F1-score) that naturally scale between 0 and 1, where higher values indicate better performance. This design allows for meaningful averaging of results across benchmarks. 

For the grammar correction benchmark, we employed a custom metric based on the Levenshtein distance, defined as $\frac{1}{1 + \text{Levenshtein distance}}$, which also produces values in the range \([0, 1]\) and is therefore compatible with the overall evaluation framework.

Across all experiments, we applied greedy decoding and then performed task-specific answer post-processing, with the exception of multiple-choice tasks for base models, which were evaluated using normalized likelihoods over the competing LLM-generated continuations.

\subsection{Base Models}

\begin{table*}[t]
\centering
\small
\begin{tabularx}{\textwidth}{
  l@{\hskip 1pt} >{\centering\arraybackslash}X@{\hskip 1pt} *{7}{>{\centering\arraybackslash}X@{\hskip 1pt}}
}
\toprule
\textbf{Model} & \textbf{Mean} & \textbf{Exams} & \textbf{Trivia} & \textbf{Declension} & \textbf{Words} & \textbf{Grammar} & \textbf{News} & \textbf{Speakers} \\
& Score & Accuracy & Accuracy & Accuracy & Accuracy & Leven-shtein* & ROUGE-L & F1-score \\
\midrule
Llama 3.1 70B & \textbf{0.450} & 0.667 & \textbf{0.615} & \textbf{0.586} & \textbf{0.379} & 0.291 & 0.160 & 0.550 \\
Gemma 2 27B & 0.419 & \textbf{0.693} & 0.493 & 0.505 & 0.354 & \textbf{0.306} & \textbf{0.164} & n/a \\
Qwen2 72B & 0.333 & 0.659 & 0.456 & 0.306 & 0.188 & 0.257 & 0.134 & \textbf{0.591} \\
Mistral Nemo Base (12B) & 0.322 & 0.487 & 0.418 & 0.416 & 0.237 & 0.245 & 0.130 & 0.476 \\
Llama 3.1 8B & 0.268 & 0.405 & 0.386 & 0.268 & 0.182 & 0.242 & 0.126 & 0.522 \\
Gemma 2 9B & 0.345 & 0.616 & 0.380 & 0.396 & 0.259 & 0.263 & 0.157 & n/a \\
\bottomrule
\end{tabularx}
\caption{Base LLM performances across 7 Estonian-language tasks with the metrics used for each task ordered by the average score in descending order. The ``Speakers'' task was excluded from the average, as Gemma models failed on this task due to too small context length.}
\label{tab:leaderboard-base}
\end{table*}

Table \ref{tab:leaderboard-base} shows the results of base models across the benchmarks, together wth the average score. 
We observe that the ranking of models varies considerably across the tasks. Overall, Gemma-2 27B and Llama-3 70B consistently achieved the best results, with the exception of the trivia benchmark, where Llama-3 70B significantly outperformed other models.

\subsection{Instruction-Tuned Models}

\begin{table*}[t]
\centering
\small
\begin{tabularx}{\textwidth}{
  l@{\hskip 0pt} >{\centering\arraybackslash}X@{\hskip 1pt} *{7}{>{\centering\arraybackslash}X@{\hskip 0pt}}
}
\toprule
\textbf{Model} & \textbf{Mean} & \textbf{Exams} & \textbf{Trivia} & \textbf{Decl.} & \textbf{Words} & \textbf{Grammar} & \textbf{News} & \textbf{Speakers} \\
& Score & Accuracy & Accuracy & Accuracy & Accuracy & Leven-shtein* & ROUGE-L & F1-score \\
\midrule
\textit{Proprietary models} \\
\midrule
Gemini 2.5 Pro & \textbf{0.638} & 0.897 & \textbf{0.889} & \textbf{0.955} & \textbf{0.631} & 0.321 & \textbf{0.138} & \textbf{0.820} \\
GPT-5  & 0.587 & 0.875 & 0.755 & 0.934 & 0.529 & 0.313 & 0.114 & 0.806 \\
Claude Sonnet 3.7 & 0.626 & \textbf{0.905} & 0.816 & 0.937 & 0.570 & 0.391 & 0.135 & 0.785 \\
Claude Sonnet 4 & 0.526 & 0.876 & 0.638 & 0.822 & 0.450 & 0.236 & 0.131 & 0.776 \\
GPT-4o & 0.616 & 0.900 & 0.770 & 0.948 & 0.555 & \textbf{0.395} & 0.127 & 0.724 \\
Gemini 2.0 Flash & 0.569 & 0.900 & 0.635 & 0.932 & 0.540 & 0.282 & 0.124 & 0.748 \\
GPT-4o Mini & 0.438 & 0.784 & 0.525 & 0.554 & 0.331 & 0.321 & 0.115 & 0.651 \\
\midrule
\textit{Large open models} \\
\midrule
Kimi-K2-0905 (1000B-A32B) & \textbf{0.533} & \textbf{0.880} & 0.598 & \textbf{0.876} & 0.435 & 0.270 & \textbf{0.140} & \textbf{0.783} \\
Llama-4 Maverick (400B-A17B) & 0.510 & 0.841 & 0.549 & 0.828 & \textbf{0.453} & 0.261 & 0.131 & 0.656 \\
Llama-3.1 405B Instruct & 0.475 & 0.817 & 0.530 & 0.680 & 0.410 & \textbf{0.286} & 0.129 & 0.702 \\
Deepseek V3 (671B-A37B) & 0.482 & 0.820 & \textbf{0.619} & 0.631 & 0.452 & 0.256 & 0.113 & 0.668 \\
Qwen3-235B-A22B-2507 & 0.422 & 0.829 & 0.510 & 0.555 & 0.324 & 0.207 & 0.108 & 0.713 \\
\midrule
\textit{Medium-sized open models} \\
\midrule
Llama-4 Scout (109B-A17B) & \textbf{0.488} & \textbf{0.824} & 0.538 & \textbf{0.766} & \textbf{0.407} & \textbf{0.270} & 0.125 & 0.622 \\
Llama-3.1 70B Instruct & 0.401 & 0.760 & \textbf{0.549} & 0.464 & 0.306 & 0.199 & \textbf{0.129} & \textbf{0.674} \\
Gemma-3 27B Instruct & 0.386 & 0.756 & 0.459 & 0.404 & 0.356 & 0.226 & 0.118 & 0.471 \\
Qwen3-Next-80B-A3B-Instruct & 0.356 & 0.776 & 0.449 & 0.362 & 0.246 & 0.195 & 0.109 & 0.585 \\
Gemma-2-27B-It & 0.365 & 0.760 & 0.408 & 0.406 & 0.266 & 0.223 & 0.128 & n/a \\
Qwen2-72B-Instruct & 0.280 & 0.739 & 0.443 & 0.177 & 0.152 & 0.060 & 0.112 & 0.591 \\
GPT-OSS-120B & 0.372 & 0.777 & 0.540 & 0.356 & 0.290 & 0.194 & 0.074 & 0.000 \\
\midrule
\textit{Small open models} \\
\midrule
Llama-3.1 8B EstLLM-0825 & \textbf{0.454} & 0.575 & \textbf{0.586} & \textbf{0.811} & 0.327 & \textbf{0.275} & \textbf{0.152} & 0.452 \\
EuroLLM-9B-Instruct & 0.409 & 0.610 & 0.508 & 0.644 & \textbf{0.338} & 0.248 & 0.104 & 0.038 \\
Gemma-3-12b-It & 0.354 & \textbf{0.736} & 0.426 & 0.356 & 0.289 & 0.197 & 0.119 & 0.000 \\
Mistral Nemo Instruct (12B) & 0.271 & 0.570 & 0.386 & 0.274 & 0.177 & 0.187 & 0.034 & 0.410 \\
Gemma-2-9B-It & 0.290 & 0.556 & 0.386 & 0.308 & 0.178 & 0.182 & 0.128 & n/a \\
Llama-3.1 8B Instruct & 0.244 & 0.542 & 0.385 & 0.099 & 0.132 & 0.189 & 0.119 & 0.505 \\
Qwen3-4B-Instruct-2507 & 0.212 & 0.553 & 0.329 & 0.056 & 0.090 & 0.136 & 0.105 & \textbf{0.540} \\
\bottomrule
\end{tabularx}
\caption{Instruction-tuned LLM performances across the tasks. The ``Speakers'' task was excluded from the average score computation, as some models completely failed on it. The scores between base LLMs and instruction-tuned LLMs are not directly comparable, as the tasks use different instructions and different few-shot setups.}
\label{tab:leaderboard-chat}
\end{table*}

To allow easier analysis, instruction-tuned models were divided into four categories: proprietary models from Google (Gemini), OpenAI (GPT) and Anthropic (Claude), and open models divided into small (up to 12B parameters), medium (up to 235B parameters) and large sizes. We acknowledge that this division is somewhat arbitrary.

Most recent proprietary models are \textit{reasoning models}, meaning they generate internal ``thinking tokens'' that are typically hidden from the end user, before producing the final response. In our experiments, these models were configured for minimal reasoning by setting the model parameter \texttt{reasoning\_effort=minimal}.

Table \ref{tab:leaderboard-chat} depicts the normalized results of the instruction-tuned models across the tasks. Note that the large and medium sized open models were evaluated using different API providers available through OpenRouter\footnote{During the benchmarking runs, we noticed that the quality of large open models varied dramatically, based on the actual API provider that was used by OpenRouter to fulfill the request. In order to mitigate this issue, we configured OpenRouter not to use those API providers: NovitaAI, Baseten, Together, Fireworks, SambaNova, AtlasCloud, Hyperbolic.}, while the small models were evaluated on our own server.

Among the tested models, Gemini 2.5 Pro achieved the highest average performance and the best scores in five individual tasks. Claude Sonnet 3.7 and GPT-4o each obtained highest ranked result in one benchmark. 

While the large open models lag behind the highest-ranked proprietary models in all categories, the difference is surprisingly small. For example, the best open model -- Kimi-K2 from Moonshot AI -- equals Claude Sonnet 4 in the average score (albeit it must be acknowledged that Claude Sonnet 4 resulted in noticeable regression in its Estonian abilities, compared to Claude Sonnet 3.7). As expected, the results are highly correlated with models' size, especially in the Exams and Trivia benchmarks that largely rely on factual knowledge. For tasks that expect more linguistic abilities (Declension, Words, Grammar), there is higher variety among similarly-sized models, probably resulting from different pretraining and instruction-tuning data.

The highest-ranked small model is Llama-3.1 finetuned by the EstLLM project\footnote{\url{https://huggingface.co/tartuNLP/llama-estllm-protype-0825}}, with the specific goal of improving its Estonian abilities. This indicates that finetuning using language-specific datasets is a highly useful method for this goal, as the 8B EstLLM model outperforms all other small models and all but one medium-sized models.

\subsection{Correlations between tasks}

\begin{figure}[tb]
    \includegraphics[width=\columnwidth]{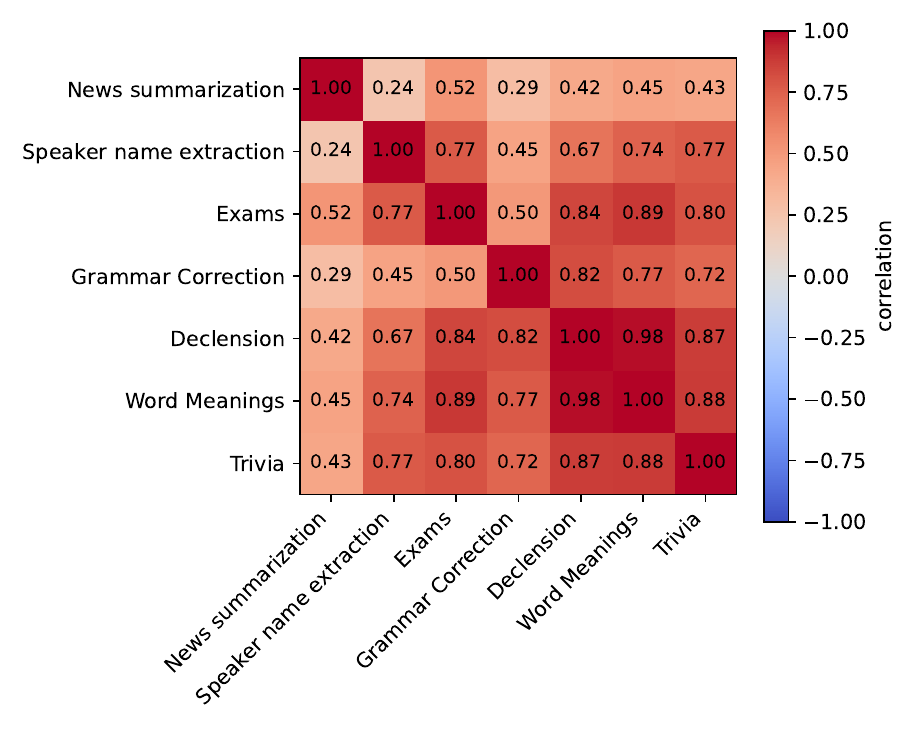}
    \caption{Pearson correlation matrix between the results of different datasets.}
    \label{fig:corr} 
\end{figure}

To explore the relationships between task types, we computed Pearson correlation coefficients across all benchmark results (Figure \ref{fig:corr}). No negative correlations were observed, indicating general consistency in model performance rankings. The strongest correlation (r = 0.98) was found between the inflection and word definition datasets, suggesting they assess highly overlapping linguistic capabilities. These tasks, along with trivia, exams, and grammar, form a tightly interrelated group, with consistently strong inter-correlations, reflecting shared demands on syntactic precision and lexical knowledge. In contrast, the news summarization dataset exhibited the weakest correlations with other tasks. This divergence implies that ROUGE-based evaluation may not reliably reflect actual model quality in text generation tasks, especially when applied to noisy or semantically variable outputs such as news content. Overall, the analysis suggests that while most benchmarks target related competencies, summarization behaves anomalously and may require improved evaluation methodologies.

\subsection{Impact of model scale}

\begin{figure}[tb] 
    \includegraphics[width=\columnwidth]{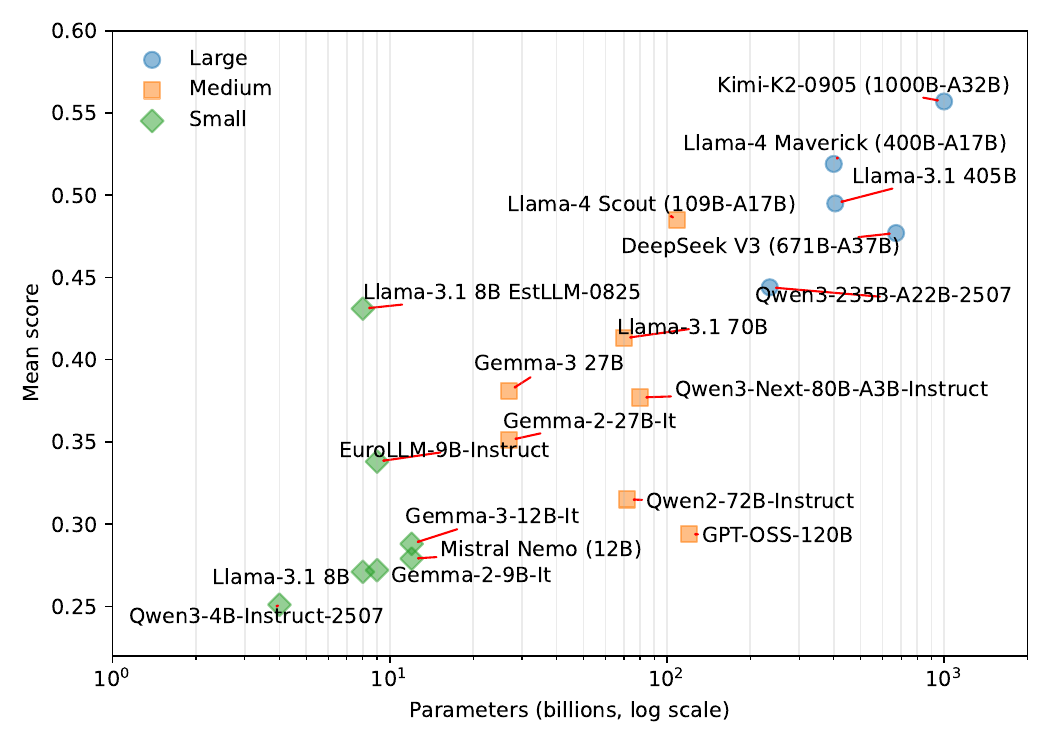}
    \caption{Model size vs. mean score (open models only). Absolute parameter size was used.}
    \label{fig:scale} 
\end{figure}

Figure~\ref{fig:scale} illustrates the relationship between model size and performance, based on the total parameter count rather than the number of active parameters, which may also influence deployment considerations. As expected, a clear positive correlation can be observed between model size and quality. Notably, the small Llama~3.1 model fine-tuned on Estonian data performs significantly above the general trendline, while some models (e.g. GPT-OSS-120B) perform below what their size would suggest.

\section{Validation}
To evaluate whether the developed Estonian-language benchmark reflects real-world usefulness, we conducted a validation study involving both human judgments and LLM-based evaluation. The goal was to assess whether model performance on benchmark tasks correlates with how helpful their outputs are perceived to be by users. This dual-validation approach provides insight into both the practical reliability of the benchmarks and the feasibility of using LLMs as automated evaluators.

\subsection{Human Evaluation}
A custom evaluation set was created using 47 questions translated from the MT-Bench dataset, spanning categories such as reasoning, writing, mathematics, general knowledge, and humanities \cite{zheng2023judgingllmasajudgemtbenchchatbot}. Culturally irrelevant or non-Estonian content was excluded. Additionally, 10 Estonian-specific questions were authored to better reflect local contexts. All questions were open-ended and designed to simulate practical LLM usage.

Responses were generated using a subset of evaluated instruction-tuned models that were available in March 2025.  Human annotators were presented pairs of anonymized model responses via a Django-based web interface and were asked choose which one was better, or indicate if they were of equal quality. Results were scored using both win rates and ELO ratings. The win rates are shown in Table \ref{tab:human_eval}. Among the models, Claude 3.7 Sonnet received the highest average score (0.75), followed closely by Deepseek V3 (0.75) and Gemini 2.0 Flash (0.70). Surprisingly, Gemma 2 27B outperformed both GPT-4o and GPT-4o Mini in human evaluations. At the lower end, Gemma 2 9B  consistently received the weakest ratings.

\begin{table}[tb]
\centering
\begin{tabular}{lr}
\toprule
\textbf{Model} & \textbf{Score} \\
\midrule
Claude 3.7 Sonnet         & 0.754 \\
Deepseek Chat V3          & 0.746 \\
Gemini 2.0 Flash 001      & 0.699 \\
Gemma 2 27B It            & 0.642 \\
Llama 3.1 70B Instruct    & 0.552 \\
GPT-4o                    & 0.545 \\
GPT-4o Mini               & 0.533 \\
Llama 3.1 405B Instruct   & 0.500 \\
Qwen2 72B Instruct        & 0.442 \\
Llama 3.1 8B Instruct     & 0.389 \\
Mistral Nemo Instruct     & 0.327 \\
Gemma 2 9B It             & 0.211 \\
\bottomrule
\end{tabular}
\caption{Model win scores based on human evaluation.}
\label{tab:human_eval}
\end{table}

To assess alignment between human preferences and benchmark task performance, we computed Pearson and Spearman correlation coefficients between normalized benchmark scores and human rating scores (Table \ref{tab:human_eval_pearson}). The correlations were generally strong or moderate, depending on the dataset. The highest correlations were observed for the Exams (Pearson r = 0.86, Spearman $\rho$ = 0.85) and Word Definitions benchmarks (r = 0.80, $\rho$ = 0.79), indicating that benchmark scores in these domains are highly predictive of perceived model quality. Lower correlations appeared in tasks such as Grammar Correction (r = 0.48) and especially News Summarization (r = 0.44), suggesting that automatic metrics like ROUGE may not capture meaningful differences in generation quality in all cases. 

\begin{table}[tb]
\centering
\begin{tabular}{lr}
\toprule
\textbf{Dataset} & \textbf{r} \\
\midrule
Exams & 0.86 \\
Trivia & 0.80 \\
Declension & 0.72 \\
Word meanings & 0.66 \\
Speaker name extraction & 0.63 \\
Grammar & 0.48 \\
News summarization & 0.44 \\
\bottomrule
\end{tabular}
\caption{Pearson correlation coefficients (r) between benchmark test and human evaluation results. }
\label{tab:human_eval_pearson}
\end{table}

\subsection{LLM-as-a-Judge}

To complement human evaluation and explore scalability, we also tested the effectiveness of using a strong LLM as an automatic evaluator. Claude 3.7 Sonnet was selected for this role due to its strong performance in both the benchmark and human evaluations\footnote{This evaluation was done in March 2025 and thus excludes newer models.}.  The model was provided with instructions to select the better of two responses (or indicate a tie), following the same protocol used for human raters. Its judgments were then compared to those of human annotators. 

The agreement between Claude's ratings and human judgments was high, with a Pearson correlation of 0.94 and Spearman correlation of 0.92 (p < 0.001), indicating a strong linear and ordinal relationship. Table \ref{tab:claude_eval} shows the results of the LLM-as-a-judge evaluation using Claude 3.7 Sonnet. The model rankings closely matched those from human annotators (Table~\ref{tab:human_eval}), with the same top three models (Deepseek Chat V3, Claude 3.7 Sonnet, and Gemini 2.0 Flash) though in a slightly different order. Interestingly, both evaluation methods ranked Llama 3.1 70B higher than the larger 405B variant. At the lower end, Llama 8B was ranked lowest by Claude, while humans rated Gemma 2 9B as the weakest. Across both evaluation methods and scoring schemes (win rate and ELO), model rankings varied by no more than one position, indicating strong agreement overall. 

\begin{table}
\centering
\begin{tabular}{lr}
\toprule
\textbf{Model} & \textbf{Score} \\
\midrule
Deepseek V3  & 0.772 \\
Gemini 2.0 Flash 001 & 0.770 \\
Claude 3.7 Sonnet & 0.759 \\
GPT-4o & 0.677 \\
Gemma 2 27B It  & 0.654 \\
GPT-4o Mini & 0.620 \\
Llama 3.1 70B Instruct  & 0.473 \\
Llama 3.1 405B Instruct  & 0.428 \\
Qwen2 72B Instruct  & 0.403 \\
Mistral Nemo Instruct  & 0.226 \\
Gemma 2 9B It  & 0.214 \\
Llama 3.1 8B Instruct  & 0.205 \\
\bottomrule
\end{tabular}
\caption{Model win scores based on LLM evaluation using Claude Sonnet as a judge.}
\label{tab:claude_eval}
\end{table}


\section{Conclusion}

We presented a comprehensive benchmark for evaluating LLMs on Estonian-language tasks. The benchmark covers seven diverse tasks addressing grammatical inflection, lexical semantics, summarization, factual knowledge, and information extraction from long texts. Our evaluation of six base models and 26 instruction-tuned models reveals substantial variation in performance across both tasks and model types. While flagship commercial models  consistently outperform open-source models, recent large open models such as Kimi-K2 achieve competitive results in several tasks. Our findings also confirm that language-specific fine-tuning of open models leads to significant improvements across all Estonian-specific capabilities.

Our validation study shows that benchmark performance correlates strongly with human judgments of practical usefulness. In particular, the benchmark based on Estonian state exams showed the highest correlation with user-rated LLM quality in open-ended question answering. Furthermore, we demonstrate that a strong LLM -- Claude~3.7~Sonnet in our case -- can serve as a reliable automatic evaluator, closely aligning with human preferences. These results suggest that robust LLM benchmarking is feasible even for low-resource languages, and that top-tier LLMs can meaningfully support scalable evaluation.

Benchmark code and datasets are publicly released to encourage further research in Estonian NLP and to support benchmarking efforts for other underrepresented languages.

\section*{Ethics Statement}

All datasets used in this work were collected and processed in accordance with Estonian and European Union data protection and copyright regulations. 

Human evaluation was conducted with informed consent from voluntary participants. No personal or identifying information was collected during the evaluation. 

We acknowledge that LLMs may reproduce social, cultural, or gender biases present in training data. While the benchmark aims to measure general and language-specific capabilities of LLMs, not their social alignment, we encourage responsible interpretation of results and further study of fairness and bias in Estonian-language models.

\section*{Code and data availability}

Benchmarks source code and the individual datasets are available in the following repositories:
\begin{itemize}
    \item Benchmark code: \url{https://github.com/taltechnlp/lm-eval-harness-tasks-estonian}
    \item Estonian National Exam dataset: \url{https://huggingface.co/datasets/TalTechNLP/exam_et}
    \item Trivia dataset: \url{https://huggingface.co/datasets/TalTechNLP/trivia_et_verified}
    \item Declension dataset: \url{https://huggingface.co/datasets/TalTechNLP/inflection_et}
    \item Word meanings dataset:  \url{https://huggingface.co/datasets/TalTechNLP/word_meanings_et}
    \item Grammar dataset:  \url{https://huggingface.co/datasets/TalTechNLP/grammar_et}
    \item ERRNews dataset: \url{https://huggingface.co/datasets/TalTechNLP/ERRnews}
    \item Speaker name extraction dataset:  \url{https://huggingface.co/datasets/TalTechNLP/paevakaja_speakers}
\end{itemize}

\section*{Acknowledgments}

This research was supported by the Estonian Centre of Excellence in AI (EXAI), National Program for Estonian Language Technology Program (project EKTB104), both funded by the Estonian Ministry of Education and Research, and by the Estonian Language Data Research Infrastructure (KeTA).

\section*{Bibliographical References} 

\bibliographystyle{lrec2026-natbib}
\bibliography{refs.rebiber}


\end{document}